\pdfoutput=1

\documentclass[nohyperref]{article}

\usepackage{microtype}
\usepackage{graphicx}
\usepackage{subfigure}
\usepackage{booktabs} 

\usepackage{hyperref}

\usepackage[accepted]{icml2022}

\usepackage{amsmath}
\usepackage{amssymb}
\usepackage{mathtools}
\usepackage{amsthm}

\usepackage[capitalize,noabbrev]{cleveref}

\theoremstyle{plain}

\theoremstyle{definition}

\theoremstyle{remark}

\usepackage[textsize=tiny]{todonotes}

\newcommand{\calA}{\mathcal{A}}

\newcommand{\calD}{\mathcal{D}}

\newcommand{\calS}{\mathcal{S}}
\newcommand{\calT}{\mathcal{T}}

\DeclareMathOperator*{\argmax}{argmax}
\newcommand*{\argmaxl}{\argmax\limits}
\usepackage{adjustbox}
\usepackage{multirow}

\icmltitlerunning{Switch Trajectory Transformer with Distributional Value Approximation}

\begin{document}

\twocolumn[
\icmltitle{
Switch Trajectory Transformer with Distributional Value Approximation 
\\for Multi-Task Reinforcement Learning

}




\begin{icmlauthorlist}
\icmlauthor{Qinjie Lin}{nw,zeb}
\icmlauthor{Han Liu}{nw}
\icmlauthor{Biswa Sengupta}{zeb}
\end{icmlauthorlist}

\icmlaffiliation{nw}{Department of Computer Science, 
Northwestern University, USA}
\icmlaffiliation{zeb}{Zebra Technologies, London, UK}
\icmlcorrespondingauthor{Qinjie Lin}{qinjielin2018@u.northwestern.edu}
\icmlcorrespondingauthor{Biswa Sengupta}{biswa.sengupta@zebra.com}

\icmlkeywords{multi-agent, Transformers}

\vskip 0.3in
]



\printAffiliationsAndNotice{}  

\begin{abstract}
We propose SwitchTT, a multi-task extension to Trajectory Transformer but enhanced with two  striking features: 
(i) exploiting a sparsely activated model to 
reduce computation cost in multi-task offline model learning 
and 
(ii) adopting a distributional trajectory value estimator that improves policy performance, especially in sparse reward settings. 
These two enhancements make SwitchTT suitable for solving multi-task offline reinforcement learning problems, where model capacity is critical for absorbing the vast quantities of knowledge available in the multi-task dataset. 
More specifically, SwitchTT exploits switch transformer model architecture for multi-task policy learning, allowing us to improve model capacity without proportional computation cost. 
Also, SwitchTT approximates the distribution rather than the expectation of trajectory value, mitigating the effects of the Monte-Carlo Value estimator suffering from poor sample complexity, especially in the sparse-reward setting. 
We evaluate our method using the suite of ten sparse-reward tasks from the gym-mini-grid environment.
We show an improvement of 10\% over Trajectory Transformer across 10-task learning and
obtain up to 90\% increase in offline model training speed.
Our results also demonstrate the advantage of the switch transformer model for absorbing expert knowledge and the importance of value distribution in evaluating the trajectory. 
\end{abstract}

\section{Introduction}
\label{intro}

This paper studies the problem of multi-task offline reinforcement learning (RL). 
We first define a multi-task offline RL problem as learning a single policy that solves multiple tasks from previously collected data without online interaction with the environment. For example, suppose we want grocery robots to acquire a range of different behaviours (e.g. lift cans, pick up bowls and open closet). In that case, it is more practical to learn an extensive repertoire of behaviours using all previously collected datasets rather than learning each skill from scratch. 

\begin{figure}[t!]
    \centering
    \vspace{0.1in}
    \includegraphics[width=0.48\textwidth]{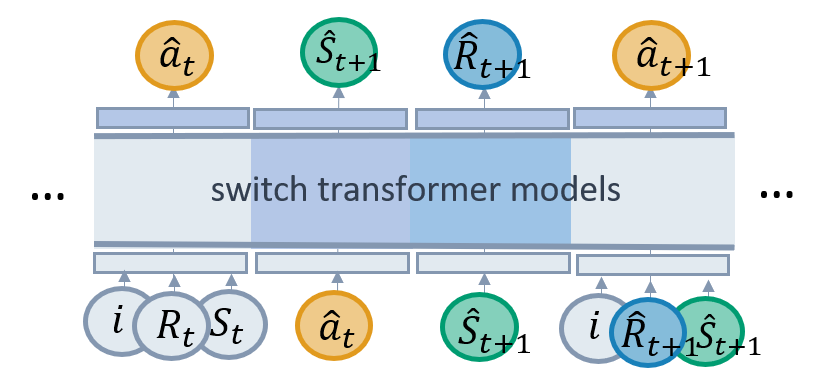}
    \vspace{-0.2in}
    \caption{
    Switch Trajectory Transformer Architecture. Switch transformer models contain action transformer, dynamics transformer and return transformer. Task id, returns, states are fed into embedding layer and consequently in to the action transformer. The model predicts the following action and then provides it to the dynamics model for predicting the next state. The return transformer uses this information and predicts return; this process is repeated until the designed horizon length. 
    }
    \vspace{-0.2in}
    \label{fig:method-arch}
\end{figure}

The large diversity of datasets collected in various tasks brings difficulty for traditional multi-task offline RL methods \cite{yu2021conservative, yu2021data, kalashnikov2021scaling}.
Specifically, these methods emphasize transferring skill knowledge across related tasks and developing a sharing experience across different tasks. Such a data-sharing strategy makes learnt multi-task policy sensitive to data distribution differences and relationships among tasks. The inherent conflict from task differences can harm the policy of at least some of the tasks, particularly when model parameters are shared among all tasks.  



Recent offline RL works like Decision Transformer \cite{chen2021decision}, and Trajectory Transformer \cite{janner2021reinforcement}, abstracting RL as sequence modelling, demonstrate the capability of turning large datasets into powerful decision-making engines. Such modelling design benefits multi-task RL problem by serving a high-capacity model for handling task differences and absorbing vast knowledge in the collected diverse dataset, and also makes it possible for multi-task RL methods to adopt the associated advances \cite{fedus2021switch} in language modelling problems. 

However, adopting such high-capacity sequential models for solving the multi-task RL problem poses three significant algorithmic challenges. The first is high computation cost and tall model capacity, which is critical for absorbing the vast knowledge available in a large heterogeneous dataset. The second one is sharing the same policy parameters across different tasks causing degraded performance over simple single-task training. 
The third is poor performance caused by the Monte Carlo value estimator, especially in sparse reward settings. Monte Carlo estimator suffers from poor sample complexity when online data collection is not allowed, and it is  uninformative to guide the beam-search-based planning procedure. 


To handle these challenges, we propose SwitchTT (Switch Trajectory Transformer), a multi-task extension to Trajectory Transformer but enhanced with two striking features. 
First, unlike Trajectory Transformer and other traditional multi-task RL methods, reusing the same parameter for all input data, our method exploits a sparsely activated model for multi-task offline model training, with the sparsity coming from selecting different model parameters for each incoming example. 
Such a model allows us to perform efficient computation in high-capacity neural networks and improve parameter sharing in multi-task learning.
Second, SwitchTT develops a trajectory-based distributional value estimator for learning the value distribution of trajectory instead of expected value, as illustrated in Figure \ref{fig:rtg-overview}. Such a distributional estimator enables us to measure and utilize uncertainty around the reward, thus mitigating the effects of the Monte-Carlo Value estimator suffering from poor sample complexity, especially in sparse-reward setting, leading to better value estimate in an offline setting. 

Inspired by the Trajectory Transformer, which abstracts offline RL as a sequence modelling problem, 
our method tackles multi-task RL with the tool of sequence modelling, utilizing the switch transformer model to model distributions over multi-task trajectories and applying beam search for planning action with the highest reward. 
A high-level overview of SwitchTT is illustrated in Figure \ref{fig:method-arch}. Specifically, we first utilize switch transformer models to train decision, return-to-go (RTG), and dynamics models, which replaces the standard feed-forward layer in the transformer with simplified Mixture of Experts (MoE) layers. The MoEs layer contains a set of expert networks and a gating network, which takes as an input an observation representation and then routes it to the best-determined expert network, producing the corresponding output. Under this MoEs layer, we view each expert network as a task learner and gating network as a router that routes the task input to the corresponding expert. Secondly, we predict trajectories based on the learned model and develop a distributional value estimator to evaluate the predicted courses and select the one with the highest reward. We will provide more details in Section \ref{sec:mth}. 


This paper has two major contributions: (i) we exploit a sparsely activated model for multi-task model learning, which reduces the computation cost and improves multi-task learning performance. (ii) we develop a trajectory-based distributional value estimator to learn a better value estimator, improving offline reinforcement learning performance. The rest of the paper is organized as below. Section \ref{sec:pre} introduces preliminaries. Section \ref{sec:mth} describes the implementation of SwitchTT. Section \ref{sec:exp} presents detailed experiment results to demonstrate advantage of SwtichTT. Section \ref{sec:related} introduces the related works and Section \ref{sec:concl} concludes the paper with more discussion.

\section{Preliminaries}
\label{sec:pre}

\paragraph{Offline Reinforcement Learning} 

Here, we define the essential reinforcement learning (RL) concepts, following standard textbook definitions \cite{sutton2018reinforcement}. Reinforcement learning addresses the problem of learning to control a dynamical system in a general sense. The dynamical system is fully defined by a Markov decision process (MDP).
The MDP is defined by the tuple \( {M} =(\calS, \calA, r, P, \gamma, \rho_0 )\), where 
\(\calS\) is the state space, \(\calA\) is the continuous action space, \(r\colon\calS \times \calA \to \mathbb{R} \) is the reward function and \({\mathcal{\rho}_0}\) represents the initial state distribution. 
\(P(s'\mid s,a)\) represents the transition probabilities, which specifies the probabilities of transition from the state \(s\) to \(s'\) under the action \(a\). 
A trajectory is made up of a sequence of states, actions, and rewards: 
$ \tau = (s_0, a_0, r_0, s_1, a_1, r_1, . . . , s_T , a_T , r_T )$. The return of a trajectory at timestep $t$, $R_t=\sum_{t'=t}^T{r_{t'}}$
is the sum of future rewards from that timestep. The goal of RL is to find an optimal policy that maximizes the expected return $\mathbb{E}[\sum_{t=1}^T{r_{t}}]$ in a MDP. 
Different than online RL, which involves iteratively collecting experience by interacting with the environment, offline RL learns the optimal policy on fixed limited dataset $\calD = \{(s_j.a_j.s_j',r_j)\}_{j=1}^{N}$, consisting of trajectory rollouts of arbitrary policies. Online interaction with the environment is too expensive and time-consuming for some systems, leading to the advantage of offline RL, which requires no additional interaction. But this setting also poses major challenges for offline RL: methods often cannot learn effectively from entire offline data without any additional on-policy interaction. Also, the exploration ability of the agent is outside the scope of such methods.

\paragraph{Multi-Task Reinforcement Learning}

The goal of multi-task RL is to find a optimal policy that maximizes expected return in multi-task Markov Decision Process (MDP), defined as \( {M} =(\calS, \calA, \{{r_i}\}_{i=1}^N, P, \gamma, \rho_0 )\), where $\{r_i\}_{i=1}^N$ is a finite set of task and others follow the definition in the offline RL subsection. Each task $i$ represents a different reward function $r_i$ but share the dynamics $P$. 
In this work, we focus on multi-task offline RL setting, aiming to find a policy $\pi(a|s)$ that maximizes expected return over all the task: \(
\pi^*(a|s) = argmax_{\pi}\mathbb{E}_{i\sim[N]}\mathbb{E}_{\pi}[\sum_{t=1}^T{r_i(s_t,a_t)}] \), given a dataset $\calD = \cup_{i=1}^{N}\calD_i$ where $\calD_i$ consists of experiences from task $i$.

\paragraph{Trajectory Transformer}

Trajectory Transformer \cite{janner2021reinforcement} formulates offline RL as generic sequence modeling problem. The core of this approach is to use a Transformer architecture to model distributions over trajectories in the offline dataset and repurposing beam search as a planning algorithm to find the optimal action. Specifically, Trajectory Transformer augment each transition in the trajectory $\tau$ with reward-to-go $R_t=\sum_{t'=t}^T\gamma^{t'-t}r_t$, then discretize a trajectory $\tau$ with N-dimension states and M-dimensional actions into sequence of length $T(N+M+1)$: $\tau = (...,s_t^1,s_t^2,...,s_t^N,a_t^1,a_t^2,...,a_t^M,r_t,...), t=0...T$ . To model distribution over such trajectories, they mirror a smaller-scale GPT \cite{radford2018improving} architecture, parameterized as $\theta$ and induced conditional probabilities as $P_{\theta}$, maximising the following objective:

\vspace{-0.2in}
\begin{equation}
  \begin{multlined}
    L(\tau) = \sum_{t=0}^{T}(
    \sum_{i=1}^{N}logP_\theta(s_t{i}|s_t^{<i}, \tau_{<t}) + \\
      logP_\theta(s_t{i}|a_t,s_t, \tau_{<t}) 
    + \sum_{j=1}^{N}logP_\theta(a_t{i}|s_t^{<i}, \tau_{<t}) 
    )
  \end{multlined}
\end{equation}
\vspace{-0.1in}

Then, beam search uses past trajectory as the input of the trained model and greedily selects the predicted trajectory ${\tau}$ with the highest reward. Other details of Trajectory Transformer are referred to in the original paper \cite{janner2021reinforcement}. 
In this work, we extend the trajectory transformer with two enhanced features to solve multi-task RL. The first is to exploit switch transformer instead of naive transformer architecture, and the second is to adopt a distributional value estimator to guide the beam search. We refer to more details to Section \ref{sec:mth}.

\paragraph{Switch Transformer}

Switch Transformer is a sparsely activated model designed to maximize the parameter count of a Transformer model in a simple and computationally efficient way. The critical difference is that instead of containing a single feed-forward neural network (FFN) in the original transformer, each switch layer has multiple FFNs known as an expert.
More specifically, the switch layer consists of a set of n ``expert networks" $ E_1, \cdot\cdot\cdot, E_n $ and a ``gating network" $G$, whose output is a sparse n-dimensional vector. The experts are themselves neural networks, each with their parameters. This layer takes a token representation $x$ as an input and then routes it to the best determined top-$k$ experts, selected from a set of ${E_i(x)}_{i=1}^n$. The router variable $W_r$ produces logits $h(x)=W_r \cdot x$ which are normalized via a softmax distribution over the available $n$ experts at that layer. The gate-value for expert $i$ is given by:
\vspace{-0.05in}
\begin{equation}
p_i(x) = \frac{e^{h(x)_i}}{\sum_j^n{e^{h(x)_j}}}.
\end{equation}
\vspace{-0.15in}

The top-$k$ gate values are selected for routing the token $x$. If $\calT$ is the set of selected top-$k$ indices then the output computation of the layer is the linearly weighted combination of each expert’s computation on the token by the gate value, 
\begin{equation}
y=\sum_{i\in \calT}{p_i(x)E_i(x)}.
\end{equation}
\vspace{-0.15in}

Instead of routing to $k$ experts, we route input tokens to only a single expert. Switch Transformer shows this simplification preserves model quality, reduces routing computation and performs better. 
The benefit of the switch layer for solving reinforcement learning problems is two-fold: (1) Each Expert is considered a task expert, such as  lifting cans, picking up bowls, and opening a closet. In this way, we can combine multiple expert knowledge inside a single policy to solve numerous tasks. (2) The gating network is considered a switch strategy, which measures the confidence of each expert and chooses the expert with the highest confidence to solve each task.

\section{Method}
\label{sec:mth}

\begin{figure*}[h!]
    \centering
    \includegraphics[width=0.9\textwidth]{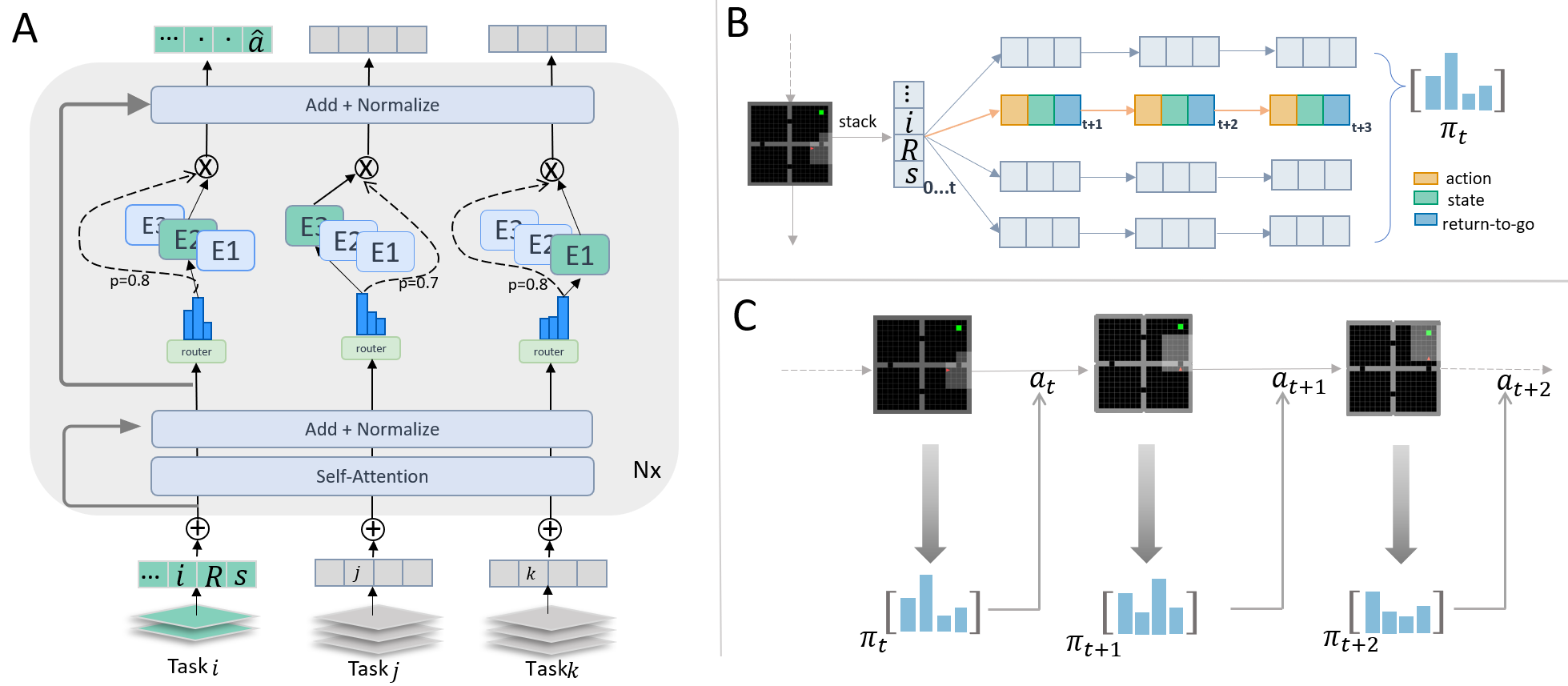}
    \vspace{-0.15in}
    \caption{
    Training, planning, and action planner in the SwitchTT: (A) Represents the action transformer model training process. Here we diagram three task tokens being routed across different experts. At first, trajectories in the collected episodes are transformed into trajectory representation by adding task id and calculating return-to-go. Then it is fed as a token into a switch transformer model, which replaces the dense feed-forward network (FFN) layer present in the transformer with a sparse Switch FFN layer. The switch layer returns the output of selected experts multiplied by the router gate value. (B) is the planning process. Firstly all visited state, return-to-go,  action and task id  are stacked into a sequence $\{i,R_0,s_0,a_0,...R_t,s_t\}$. Then the sequence is fed into switch transformer models to generate predicted four pairs of $\{a_t,s_{t+1},R_{t+1}\}$. Stack again and generate only one pair until the designed length. Then this generates values of different chosen actions at the first generation. (C) The action planner. The planner is performed at each timestep as described in B. At each timestep, the highest column value action is sampled and executed to the environment.  
    }
    \vspace{-0.05in}
    \label{fig:method-overview}
\end{figure*}
\vspace{-0.1in}

This section presents SwitchTT in detail, a multi-task extension of the Trajectory Transformer. The details of the method is illustrated in Figure \ref{fig:method-overview}. Following Trajectory Transformer, SwitchTT model multi-task offline RL problem as sequence modelling problem and divides the algorithm into three phases (data collection, offline model training and planning). Since our model and planning strategy is nearly identical to the Trajectory Transformer, we briefly describe three phases and emphasize the critical difference: the model architecture of the switch transformer model and implementation of the distributional value estimator.

\subsection{Data Collection}
In the data collection phase, we collect a combined dataset $\calD=\cup_{i=1}^{N}\calD_i$ in task $1..N$ and transform the dataset into a sequential representation. Specifically, instead of using certain-level expert policy to interact with the environment, we utilize the online RL algorithm PPO \cite{schulman2017proximal} to solve each task and collect the replay dataset $\calD_i$. This adds exploratory trajectory into the dataset and increases the diversity of trajectory inside the dataset.  
Then, trajectories from different tasks are combined into a single dataset $\calD$. Each trajectory is transformed into the trajectory representation following Decision Transformer \cite{chen2021decision}. Instead of feeding the rewards directly, we feed the model with the returns-to-go $\hat{R_t}=\sum_{t'=t}^T{r_{t'}}$. This leads to the following trajectory representation of task $i$:
\begin{equation}
\tau = (i,\hat{R_0}, s_0, a_0, i, \hat{R_1}, s_1, a_1, . . . , i,\hat{R_T}, s_T , a_T). 
\end{equation}
At test time, we feed the target return and first state $(R_0,s_0)$ into the model and then get the desired action $a_0$ from the planning phase, depicted in the latter subsection. Here, $R_0$ represents the desired performance, usually the maximized cumulative reward of the task. $s_0$ is the first observed state from the environment. After executing the action, we receive $(r_0,s1)$ from the environment and decrease the target return via equation $R_1=R_0-r_0$. Then we feed the current trajectory to the model and repeat until the episode terminates.

\subsection{Training Phase}

In the training phase, we train offline models modelling the trajectory distribution in the dataset $\calD$ and use them for the planning phase.
Instead of modelling the trajectory inside a single model like Trajectory Transformer, we train three separate models: (i) decision transformer $\theta$, modelling the joint distribution of the states, actions, and rewards sequence. (ii) dynamics transformer $f$, predicting state over past states and actions. (iii) return-to-go (RTG) transformer $\phi$, a trajectory-based distributional value estimator, predicting value distribution of state-action trajectory. Here, the training procedure of each transformer is referred to in  \citet{chen2021decision}. The training procedure of the three models is the same as the original Transformer model. The objective of the  Decision Transformer model is to minimize the cross-entropy loss between predicted and true actions. We define the loss as a cross-entropy loss between predicted and true states for the Dynamics Transformer model. We also use switch transformer model architecture instead of transformer model and describe the distributional value estimator in detail. 

We feed the last K timesteps into the decision transformer for a total of 4K tokens. For the dynamics transformer and return-to-go transformer, we provide previous k timesteps into them but only 3k tokens, including task id, state and actions. The output of the Dynamics Transformer is the next-timestep state, and the output of the return-to-go transformer is the distribution of the input trajectory's future return.

\paragraph{Trajectory-based Distributional Value Estimator} 
\begin{figure}[h!]
    \centering
    \vspace{-0.1in}
    \includegraphics[width=0.45\textwidth]{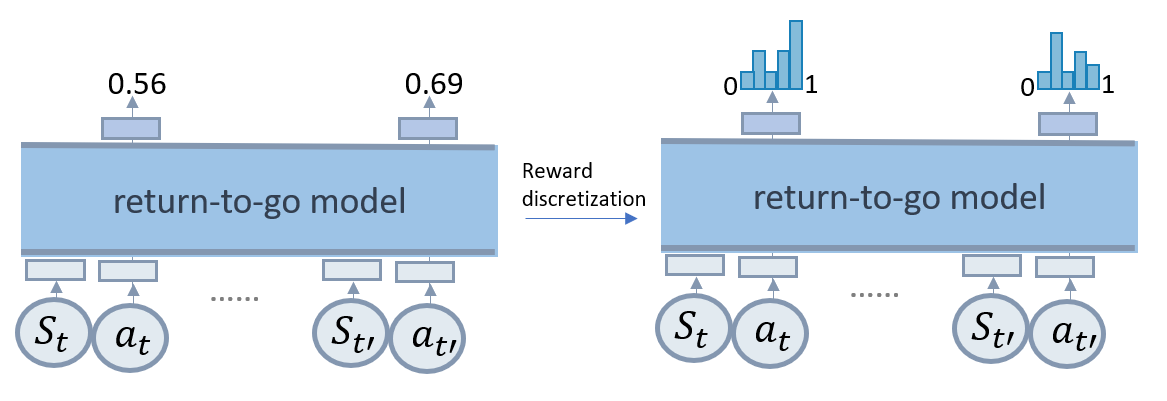}
    \vspace{-0.1in}
    \caption{Overview of return-to-go transformer. Instead of predicting the expectation value of trajectory, return-to-go transformer discretise the reward and predicts the value distribution of the trajectory. }
    \vspace{-0.25in}
    \label{fig:rtg-overview}
\end{figure}

As shown in Figure \ref{fig:rtg-overview}, return-to-go transformer is a trajectory-based distributional value estimator that estimates the value distribution of the state-action trajectory instead of the expected return. 
Here, we model the distribution using a discrete distribution (categorical distribution) parameterized by 
$N\in \mathbb{N}$ and $V_\text{min},V_\text{max}\in \mathbb{R}$.  
The support of such distribution is the set of atoms 
$\{ z_i = V_\text{min} + i	\triangle z: 0\leq i < N \}$, 
$ \triangle z = {\tfrac{V_\text{max}-V_\text{min}}{ N-1}}$ and atom $z_i$ probability is given by a parametric model $\theta: \mathcal{X} \rightarrow \mathbb{R}^N  $,  as shown in Figure \ whose input is three-step state-action trajectory $x= \{ s_t,a_t,s_{t+1},a_{t+1},s_{t+2}, a_{t+2} \}$ and output is the probability vector of atoms: 
\vspace{-0.1in}
\begin{equation}
p_i(x) = \frac{e^{\theta(x)_i}}{\sum_j^n{e^{\theta(x)_j}}}
\end{equation}
\vspace{-0.1in}

Then we project the value distribution learning onto a multiclass classification problem. Given a sample trajectory $ \tau _t = \{\hat{R_j}, s_j, a_j\}_{j=t}^{t+3}$ from dataset, we label the trajectory as class $ c =  \lfloor { \hat{R}_{t+3} - V_{\text{min}} } \rceil / \triangle z  $ and use gradient descent to find parameter $\theta$ that minimize the cross-entropy loss between labeled class and discrete value distribution. In the planning phase, the output value of given trajectory $x$ is the linearly weighted combination of each atoms' probability:

\vspace{-0.1in}
\begin{equation}
V(x)=\sum_{i=0}^{N-1}{p_i(x)z_i}.
\label{eq-value}
\end{equation}
\vspace{-0.1in}

The advantage of using distributional value learning is to learn a better value approximation. While the Monte Carlo value estimate suffers from poor sample complexity, especially in sparse-reward tasks, learning the distribution preserves the suboptimal behaviour. 
Also, by leveraging the transformer model architecture, we learn the distribution in a simple self-supervised manner instead of dynamic programming as in the Distributional RL work \cite{bellemare2017distributional}.  

\paragraph{Model Architecture} We utilize a switch transformer to learn the distribution of the combined dataset. The model architecture is extended from GPT \cite{radford2018improving} model architecture. But we replace the dense feed-forward network (FFN) layer present in the transformer with a switch layer consisting of multiple FFN layers. Figure \ref{fig:method-overview} illustrates the detail of the switch layer in the transformer models. When each token passes through this layer, it first passes through a router function and the router routes the token to a specific expert. Under the multi-task RL setting, we consider each expert as a policy model, and the router routes the observation from a task to the highest-confidence expert. In our method, we implement a mirror version of Switch Transformer architecture, and the detail of the switch layer is depicted in Switch Transformer \cite{fedus2021switch}. 

\subsection{Planning Phase}

In the planning phase, we use beam search to generate optimal action with the highest value estimation based on the imagined sequence. We summarize this phase as Switch Planner and describe it in the Algorithm \ref{alg:swtichpl}. 
The algorithm require model $(\theta,f,\phi)$ from training phase, hyperparameter candidate number $c$ and horizon length $h$, current sequence $x$ as input. Here, $c$ determines the number of imagined trajectory, $h$ define the horizon length of each imagined trajectory, $x$ is collected from past states, actions and rewards at timestep t: $x=\{R_i,s_i,a_i\}_{i=0}^{t}$. The planning phase illustrated in Figure \ref{fig:method-overview} use candidate number $c=4$ and horizon length $h=3$.

\vspace{-0.1in}
\begin{algorithm}[h!]
   \caption{SwitchTT Planner}
   \label{alg:swtichpl}
\begin{algorithmic}
   \STATE {\bfseries Input:} sequence {x}, decision model $\theta$, dynamics model $f$, reward model $\phi$, candidate number $c$, horizon length $h$.

   \STATE Initialize $\calT=\{\}$, $\calA=\{\theta(x){[i]}\}_{i=0}^c$
   
   \FOR{$i=1$ {\bfseries to} $c$}
        \STATE{
        Sample $i^{th}$ action candidate $a^i_1=\theta(x){\small{[i]}}$   
        \newline
        Initialize trajectory $\tau_i=\{x,a^i_0\}$ 
        \FOR{$j=1$ {\bfseries to} $h$}
            \STATE{  
            Predict state $s^i_j=f(\tau_i)$ 
            \newline
            Imagine trajectory $\tau_i = \tau_i \cup \{a^i_{j-1},s^i_j\} $ 
            \newline
            Sample action $a_j^i=\theta(x)[0]$ 
        }\ENDFOR 
        \newline
        Evaluate trajectory $ \calT[a^i_0] = V_{\phi}(\tau_i)$ using Eq \ref{eq-value}
        }
   \ENDFOR
   
   \STATE {\bfseries Return: $ \argmaxl_{a\in \calA} \calT[a] $ }

\end{algorithmic}
\end{algorithm}
\vspace{-0.1in}

\section{Experiment}
\label{sec:exp}

In this section, we present the experimental results of our method and compare them with baselines methods in a multi-task setting. 
In particular, our experiment aims to answer the following question: (i) How well does our method perform on multi-task learning? (2) Does switch transformer speed up the training speeds of offline model learning. (3) Does switch transformer mitigate the effect of degrading performance of multi-task learning over single-task learning? (4) Does the distributional value estimator improve multi-task performance?
 


\subsection{Experiment Design}

We evaluate our methods in 10 different tasks of the gym-mini-grid environment, including FourRooms, DoorKey, KeyCorridor, and the other seven tasks. We choose these tasks because they have sparse reward where by default, the agent only gets a positive reward when reaching the designed goal. This problem is difficult for policy learning because reward must be propagated from the beginning to the end of the episode when actions taken in the middle is skipped over. Also, these tasks contain different requirements, so the policy should be trained separately, although they have the same action and state space. For example, DoorKey is a simple sparse-reward problem; the Fourroms is a long-term credit assignment problem and KeyCorridor requires learning compositional tasks. 

We compared our methods with Trajectory Transformer (TT), Decision Transformer (DT), Behavior Cloning.
Our motivation for choosing these methods are: Our methods extend from Trajectory Transformer and are similar to Decision Trajectory, which are offline RL methods, abstracting offline RL as a sequence modelling problem. Imitation learning is similar to our methods since it also uses supervised loss for training and planning from the trained models.

\subsection{Performance in multi-task learning}

Here, we firstly investigate the improvement of SwitchTT over TT and DT     in multi-task learning. 
We collect a dataset across ten different tasks in a gym-mini-grid environment. The total amount of the dataset is 5 million timesteps combined dataset, 500k timesteps for each task. Combining all the data, we trained DT, TT and Switch TT models and evaluated the trained models on the ten tasks. We calculate the reward by running each task across 100 scenarios. 
Secondly, we study the effect of the switch layer for multi-task learning. We compare SwitchTT with other baseline methods in three learning settings, 1-task learning, 3-task learning and 10 task learning. 
We use a context length of k=30 in both experiments in all trained transformer models. In the switch layer, we use the expert number of n=3 for 3-task learning and n=8 for 10-task learning. 

\begin{figure}[h!]
    \centering
    \includegraphics[width=0.48\textwidth]{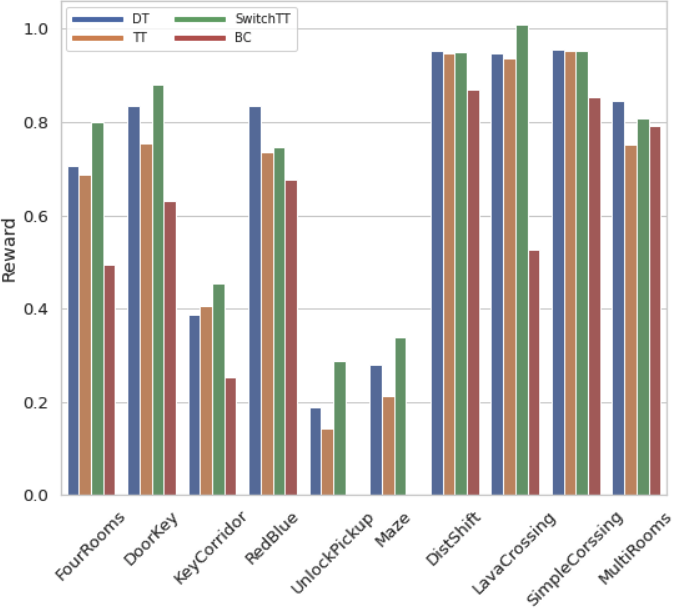}
    \vspace{-0.25in}
    \caption{ Evaluation of Switch TT across 10 tasks. DT and TT train transformer models with FFN layer on the combined 10-task datasets. SwitchTT train transformer models with Switch layer on the same dataset. SwitchTT provides an improvement over 3 baseline methods across 80\% of tasks.
    }
    \label{fig:task-performance}
\end{figure}

As shown in Figure \ref{fig:task-performance}, SwitchTT outperforms DT, TT across 80\% of tasks.  In particular, SwithchTT improves about 15\% performance over other methods in the FourRooms, DoorKey and Maze tasks, which require long-horizon planning ability. 
This highlights that the transformer models with switch layers learn a better-matched trajectory distribution in the dataset than FFN layers. Also, the distributional value estimator benefits SwitchTT by providing a more accurate value for estimating the imagined trajectories.

\begin{figure}[h!]
    \centering
    \includegraphics[width=0.30\textwidth]{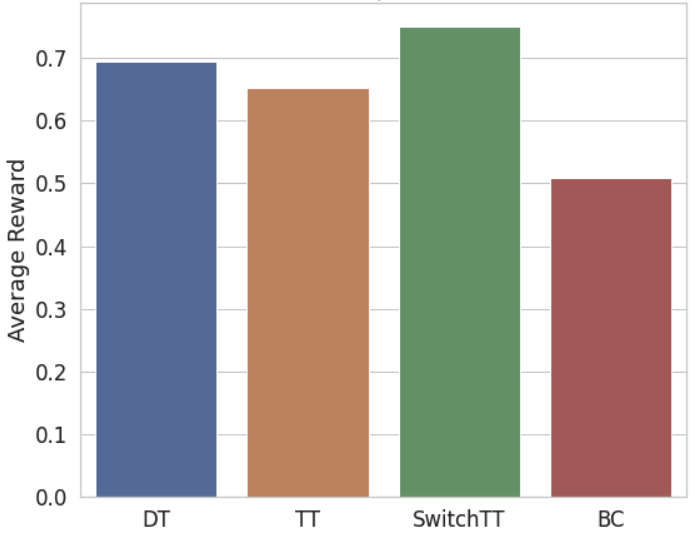}
    \vspace{-0.1in}
    \caption{
    The average reward of 10-task learning. We train offline models on the 10-task dataset, evaluate the performance on ten tasks, and report the average reward of 10 tasks in 10-task learning. 
    }
    \label{fig:avg-reward}
\vspace{-0.2in}
\end{figure}

Figure \ref{fig:avg-reward} shows that SwitchTT achieves the best result in 10-task learning. We observe that our methods improve about 10\% over our base methods TT. These highlights switch layers improve the offline model learning, and also the distributional value estimator also learns a better value in a sparse-reward setting. 

\begin{table}[h!]
\begin{adjustbox}{width=\columnwidth}
\centering
\small
\begin{tabular}{ c | c | c | c | c  } 
 \hline
\addlinespace[0.08cm]
 & Method & {1-task} & {3-task} & {10-task}  \\
\addlinespace[0.08cm]
 \hline
 
\addlinespace[0.05cm]
\multirow{6}{*}{\small {FourRooms}} 
 & SwitchTT   &          $0.716\pm0.11$ &          $0.724\pm0.08$ & \boldmath $0.800\pm0.11$ \\
 & TT         & \boldmath$0.766\pm0.09$ &          $0.658\pm0.09$ &          $0.686\pm0.09$ \\
 & TT-DV      &          $0.716\pm0.11$ &          $0.735\pm0.09$ &          $0.747\pm0.10$ \\
 & BC         &          $0.563\pm0.15$ &          $0.543\pm0.15$ &          $0.494\pm0.12$ \\
 & DT         &          $0.682\pm0.15$ & \boldmath$0.736\pm0.09$ &          $0.706\pm0.10$ \\
 & DT-Switch  &          $0.682\pm0.15$ &          $0.682\pm0.14$ &          $0.741\pm0.12$ \\
\addlinespace[0.05cm]
\hline

\addlinespace[0.1cm]
\hline

\addlinespace[0.05cm]
\multirow{6}{*}{\small {DoorKey}} 
 & SwitchTT   &          $0.899\pm0.04$ & \boldmath$0.875\pm0.07$ & \boldmath$0.880\pm0.12$ \\
 & TT         &          $0.882\pm0.04$ &          $0.855\pm0.07$ &          $0.754\pm0.08$ \\
 & TT-DV      &          $0.899\pm0.04$ &          $0.828\pm0.05$ &          $0.831\pm0.12$ \\
 & BC         &          $0.825\pm0.08$ &          $0.728\pm0.14$ &          $0.630\pm0.08$ \\
 & DT         & \boldmath$0.912\pm0.06$ &          $0.861\pm0.07$ &          $0.835\pm0.12$ \\
 & DT-Switch  &          $0.912\pm0.06$ &          $0.873\pm0.08$ &          $0.867\pm0.11$ \\
\addlinespace[0.05cm]
\hline

\addlinespace[0.1cm]
\hline

\addlinespace[0.05cm]
\multirow{6}{*}{\small {KeyCorridor}} 
 & SwitchTT   &          $0.335\pm0.08$ &          $0.383\pm0.10$ & \boldmath$0.453\pm0.10$ \\
 & TT         &          $0.096\pm0.06$ &          $0.380\pm0.08$ &          $0.404\pm0.07$ \\
 & TT-DV      &          $0.235\pm0.08$ &          $0.310\pm0.06$ &          $0.354\pm0.06$ \\
 & BC         &          $0.384\pm0.15$ &          $0.427\pm0.15$ &          $0.252\pm0.11$ \\
 & DT         &          $0.541\pm0.11$ &          $0.510\pm0.08$ &          $0.388\pm0.08$ \\
 & DT-Switch  & \boldmath$0.541\pm0.11$ & \boldmath$0.5139\pm0.14$ &          $0.428\pm0.09$ \\
\addlinespace[0.05cm]
\hline

\end{tabular}
\end{adjustbox}
\vspace{-0.1in}
\caption{The effect of Switch Layer for multi-task learning. The table reports the reward mean and variance of baseline methods in three multi-task learning settings. n-task learning in the table means trained datasets contain trajectories of n tasks. TT-DV stands for the Trajectory Transformer but is enhanced with a distributional value estimator. BC and DT-switch trains transformer models with switch layers.  }
\vspace{-0.15in}
\label{table:multi-performance}
\end{table}


We highlight three key findings from Table \ref{table:multi-performance}: 
(1) SwitchTT outperforms other baseline methods in 10-task learning. For the multi-task setting, SwitchTT can achieve the best results.
(2) DT-Switch performs better in 3 multi-task learnings than DT,  cooperating with the Switch Layer in the transformer model. DT-Switch improves the multi-task learning performance over DT using the FFN layer in the transformer model. We conclude that such a switch layer mitigates the degrading performance of multi-task learning over single-task learning.
(3) TT-DV outperforms TT in 3 multi-task learning settings. The distributional Value estimator improves the performance of the TT. We conclude that our distributional value estimator provides better value estimation for TT and improves the performance of TT.


\subsection{Computation Cost of Model Learning}

Here, we design experiments to compare the computation cost of the Switch Transformer model with the baseline model in learning multi-task dataset. 
In particular, we use PPO to collect the multi-task dataset in 10 gym-mini-grid tasks and train six models on the collected dataset. The six models are the Decision Transformer (DT) Model and DT with switch layer (DT-Switch) models with three different transformer models, large, medium and small. The Decision Transformer model is a torch-implemented mirror version of the GPT transformer model. The DT-Switch is the same model but replaces the FFN layer with the switch layer. The head, layer, embedding size of small, medium, larger are $(4,4,64)$,$(8,4,64)$ and $(8,4,128)$. The expert number in the switch layer is 4. During the training process, we plot the training loss. Then instead of plotting test loss, we test the model performance on three environments, 100 tasks for each environment. 


\begin{figure}[h!]
    \vspace{-0.05in}
    \centering
    \includegraphics[width=0.45\textwidth]{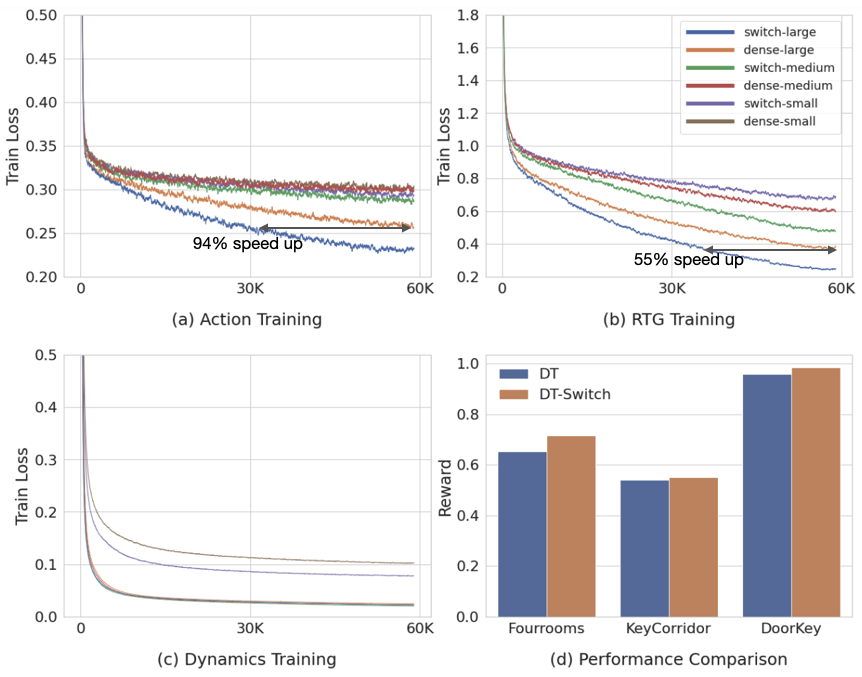}  
    \vspace{-0.15in}
    \caption{Comparison of time cost and performance between dense layer (FFN) and switch layers. Subfigure a, b, c plots the training loss of action model return-to-go model, dynamics model. Subfigure d plots the reward performance of DT and DT-Switch on three tasks.     }
    \label{fig:switch-comparison}
    \vspace{-0.10in}
\end{figure}

Figure \ref{fig:switch-comparison}.a, \ref{fig:switch-comparison}.b, \ref{fig:switch-comparison}.c shows that transformer models with switch layer always converge to a lower train loss with faster speed. This means switch layers can reduce the computation cost, including training time and parameter size for multi-task model learning. Also, Figure \ref{fig:switch-comparison} shows that DT-Switch outperforms DT on three tasks. This implies DT-switch mitigates the effect of overfitting the models and also learns a better distribution of trajectories in the multi-task dataset. This result demonstrates the advantage of such a sparsely activated layer in multi-task learning. 



\paragraph{Effect of expert number} 

\begin{table}[h!]
\begin{adjustbox}{width=\columnwidth}
\centering
\small
\begin{tabular}{ c | c | c | c  } 

 \hline
\addlinespace[0.08cm]
{Expert} & {FourRooms} & {DoorKey} & {KeyCorridor}  \\
\addlinespace[0.08cm]

 \hline
\addlinespace[0.05cm]

N=1 &          $0.7056\pm0.10$  &          $0.8348\pm0.12$  &          $0.3882\pm0.08$ \\
N=2 &          $0.6432\pm0.15$  &          $0.8612\pm0.08$  &          $0.4127\pm0.15$ \\
N=4 & \boldmath$0.7413\pm0.12$  & \boldmath$0.8667\pm0.11$  & \boldmath$0.4276\pm0.09$ \\
N=8 &          $0.6145\pm0.17$  &          $0.8339\pm0.12$  &          $0.2245\pm0.13$ \\
N=16 &         $0.6605\pm0.14$  &          $0.8344\pm0.12$  &          $0.3182\pm0.11$ \\

\addlinespace[0.05cm]
\hline
\end{tabular}
\end{adjustbox}
\vspace{-0.1in}
\caption{Effect of Expert numbers in a 10-task learning setting. We report the reward mean and variance value across 10 different seeds in 10-task learning setting. The $N$ represents the expert number of switch layer in the switch transformer models.  }
\vspace{-0.1in}
\label{table:expertNumber}
\end{table}

Table \ref{table:expertNumber} compares the performance of different expert numbers in 10-task learning. Here we report the reward mean and variance in 3 difficult tasks, FourRooms, DoorKey and KeyCorridor. The model with four experts in the switch layer performs better than others. Notably, the model with higher expert numbers 8,16 performs worse than number 4. We conclude that the model with four experts performs best in 10-task learning.

\subsection{Performance of Distributional Value Estimator }

\textbf{Effect of Atom Number}
Here, we design experiments to compare the performance of different types of return-to-go transformer models.
First, we implement four types of return-to-go transformer models in the training phase, which discretize the reward into 11 atoms, 31atoms, 51 atoms, 101 atoms. We train the SwitchTT in a 10-task learning setting and report the reward mean and variance of 3 typical but complex tasks, FourRooms, DoorKey, KeyCorridor. 

\begin{table}[h!]
\begin{adjustbox}{width=\columnwidth}
\centering
\small
\begin{tabular}{ c | c | c | c  } 

 \hline
\addlinespace[0.08cm]
{Atom} & {FourRooms} & {DoorKey} & {KeyCorridor}  \\
\addlinespace[0.08cm]

 \hline
\addlinespace[0.05cm]

N=11   &          $0.7742\pm0.07$  &          $0.5863\pm0.09$  &          $0.2295\pm0.13$ \\
N=31   & \boldmath$0.7816\pm0.10$  &          $0.7457\pm0.11$  &          $0.1122\pm0.07$ \\
N=51   &          $0.7704\pm0.09$  &          $0.7583\pm0.11$  &          $0.1568\pm0.13$ \\
N=101  &          $0.7768\pm0.08$  & \boldmath$0.8203\pm0.12$  &  \boldmath$0.2785\pm0.09$ \\   

\addlinespace[0.05cm]
\hline
\end{tabular}
\end{adjustbox}
\vspace{-0.1in}
\caption{Effect of Atom Number in the distributional value estimator. We report the reward mean and variance values across 10 different seeds in 10-task learning setting. Atom number stands for the number of discretizing the reward.  }
\vspace{-0.1in}
\label{table:dist-type}
\end{table}

In Table \ref{table:dist-type}, we study the effect of atom number in our distributional value estimator. We observe that the highest atom number has the highest performance in DoorKey and KeyCorridor tasks, which rely on an accurate value estimator. Specifically, DoorKey and KeyCorridor tasks need to consider move action and open door, pick keys, drop keys action while FourRooms task only requires move action. We conclude that increasing the atom number in the distributional value estimator improves the model performance in multi-task learning.

\textbf{Improvement over baseline}
Table \ref{table:multi-performance} reports the reward mean and variance of TT and TT-DV. We can see that the average performance of TT-DV in different multi-task settings outperforms TT. We conclude that the distributional value estimator improves the trajectory transformer performance by providing a more accurate value estimation of trajectories. 
\section{Related Works}
\label{sec:related}

This section summarizes some related works. 

\textbf{Transformer for Reinforcement Learning}
Decision Transformer \cite{chen2021decision} model Reinforcement Learning (RL) as a sequence modelling problem and matches or exceeds the performance of state-of-the-art model-free offline RL baselines. 
Based on this promising result, recent works draw upon the simplicity and scalability of the Transformer architecture to solve the reinforcement learning problem. These works can be divided into two kinds of approaches.
The first is similar to Decision Transformer, which can be viewed as model-free RL at a high level. This suite of frameworks \cite{shang2021starformer,yang2021representation} model the conditional distribution of actions given trajectory data. 
The other line of work \cite{janner2021offline,chen2022transdreamer,hafner2019dream} is viewed as model-based RL at a high level, which not only models the conditional distribution of action given trajectories but also model the state transition over the trajectory. 
The latter approach is more reliable in solving long-horizon sparse-reward tasks because the learned dynamics allow for future trajectories. The planning phase predicts the cumulative reward over a long horizon. 
Our method lies in the second line of work, which learns the state transition over trajectory data and utilizes the learned model to imagine the future trajectory.
Different from trajectory transformer, learning the value estimation by temporal difference learning, we improve the Monte Carlo value estimate by introducing  distributional value learning. Instead of estimating the expected value return, we model the value learning as a categorical distribution and utilize the transformer to learn the distribution. 

\textbf{Model-based Offline Reinforcement Learning}
Existing works have demonstrated the promise of model-based RL for offline learning. 
A command approach for model-based offline RL focuses on learning a dynamics model for uncertainty estimation and then optimizing the policy. For example, 
Model-based Offline Reinforcement learning (MOReL) firstly learns a pessimistic MDP from offline data using Gaussian dynamics model and then learns a policy for the learned MDP \cite{kidambi2020morel}; 
Model-based Offline Policy Optimization (MOPO) estimates learned model error and penalizes rewards by such error to avoid distributional shift issues \cite{yu2020mopo};
Offline Reinforcement Learning from Images with Latent Space Models (LOMPO) extends MOPO to high-dimensional visual observation spaces \cite{rafailov2021offline}. 
These approaches do not directly use the learned model to plan action sequences.
MuZero combines a tree-based search with a learned model, using the learned model directly for policy and value improvement through  online planning \cite{schrittwieser2020mastering}. MuZero Unplugged \cite{schrittwieser2021online} extends MuZero to an offline-RL scenario and achieves state-of-the-art results. 
In contrast, our method utilizes sequence modelling tools to model the distribution of trajectories in the offline dataset. Such a high-capacity sequence model architecture provides a more reliable long-horizon predictor than a conventional dynamics model. It mitigates the effect of accumulated predictive error over a long horizon.

\textbf{Multi-Task Reinforcement Learning}
Multi-task Reinforcement Learning (RL) aims to learn a single policy that efficiently solves multiple skills. Prior works have made promising result but still face three major challenges, including optimization difficulties \cite{schaul2019ray,hessel2019multi,yu2020gradient}, effective weight sharing for learning shared representations \cite{teh2017distral, espeholt2018impala, xu2020knowledge, d2019sharing, sodhani2021multi, stooke2021decoupling} and sharing data across different tasks \cite{eysenbach2020rewriting,kalashnikov2021mt,yu2021conservative}. We study the challenge of effective weight sharing for learning shared representations in the multi-task offline RL setting. Current works focus on learning shared representation across different tasks, then apply traditional RL like computing policy gradient or learning value function to solve multiple tasks. In contrast, we abstract the multi-task RL problem as a sequence modelling problem and apply a high-capacity transformer model to solve the multi-task RL problem.

\section{Conclusion}
\label{sec:concl}
We propose SwitchTT, seeking to solve multi-task reinforcement learning via an advanced transformer model. Promising experimental results show that our method outperforms other offline RL methods. Future work will consider combined advanced tree search algorithms like Monte-Carlo Tree Search to improve the performance.


\newpage
\bibliography{main}
\bibliographystyle{icml2022}

\newpage
\appendix
\onecolumn

\section{Related works}


\paragraph{Transformer} 
Transformer \citep{vaswani2017attention} has been proposed as a novel model architecture to handle sequential input data in machine translation tasks. Since then, transformer-base pretrained language models like GPT-2/3 \citep{radford2019language,brown2020language},  BERT \citep{devlin2018bert}, XLNet \citep{yang2019xlnet}, RoBERTa \citep{liu2019roberta}, T5 \citep{raffel2019exploring}, ALBERT \citep{lan2019albert}, BART \citep{lewis2019bart}  have achieved tremendous success in NLP because of their ability to learn language representations from large volumes of unlabeled text data and then transfer this knowledge to downstream tasks. 
In light of the above works, researchers are tempted to investigate the benefit of transformer models in improving reinforcement learning performance. The first line of work applies the transformer model to represent the component in standard RL algorithms, such as policy, models and value functions \citep{parisotto2020stabilizing,parisotto2021efficient}. Instead of this, the second line of work \citep{chen2021decision,janner2021reinforcement} abstracts  RL as a sequence modelling problem and efficiently utilize the existing transformer framework widely used in language modelling to solve the RL problem. Intuitively, the latter approach is more influential since it can support the possibility that the advances in sequence models can directly be applied to the RL problem without relying on the RL algorithm framework. 
Motivated by the second approach, we model multi-task reinforcement learning problem as a multi-lingual task problem in Natural Language fields, which allows us to draw upon existing advanced transformer frameworks \cite{fedus2021switch,lepikhin2020gshard}. Unlike previous works learning single task, our work makes efficient use of a high-capacity model to acquire multiple tasks inside a single model. 

\vspace{-0.1in}
\paragraph{Mixture of Experts} 
The mixture-of-experts(MoEs) approach \citep{jordan1994hierarchical, jacobs1991adaptive} was proposed more than two decades ago to divide the problem space into homogeneous regions. Recent works on MoEs can be divided into two kinds of approaches. The first is to propose different types of architecture such as SVMs \citep{collobert2002parallel}, Bayesian Methods \citep{waterhouse1996bayesian} and Gaussian Processes \citep{tresp2001mixtures}.
While the above work considers mixture-of-experts as the whole model, the second kind of work use mixture-of-expert as a part of neural network. Recent work \citep{eigen2013learning} extends MoEs to use a gating network at each layer in a multilayer network, forming a Deep Mixture of Experts. 
Based on this idea, Shazeer \citep{shazeer2017outrageously} uses MoEs as a general-purpose neural network component and significantly advances state-of-the-art results on public language modelling data sets. GShard \citep{lepikhin2020gshard} and Switch Transformer \citep{fedus2021switch} then adopt the MoEs into transformer model architecture to scale the model size efficiently. This dramatically reshaped the landscape of natural language processing research. 
It's intuitive that the latter approach efficiently scales the model capacity and improves model performance on complex problems since sub-problems in a complex problem require different expert solvers. 
Inspired by this novel architecture, we extend the MoEs to solve the problem of multi-task reinforcement learning. In our work, the MoEs layer consists of several feed-forward sub-networks, and a trainable gating network. The gating network determines a sparse combination of these experts to use for each task. The sub-networks here are considered as policy experts to solve the multi-task problem.

\section{Overview of Three models}

\begin{figure}[h!]
    \centering
    \vspace{-0.1in}
    \includegraphics[width=0.9\textwidth]{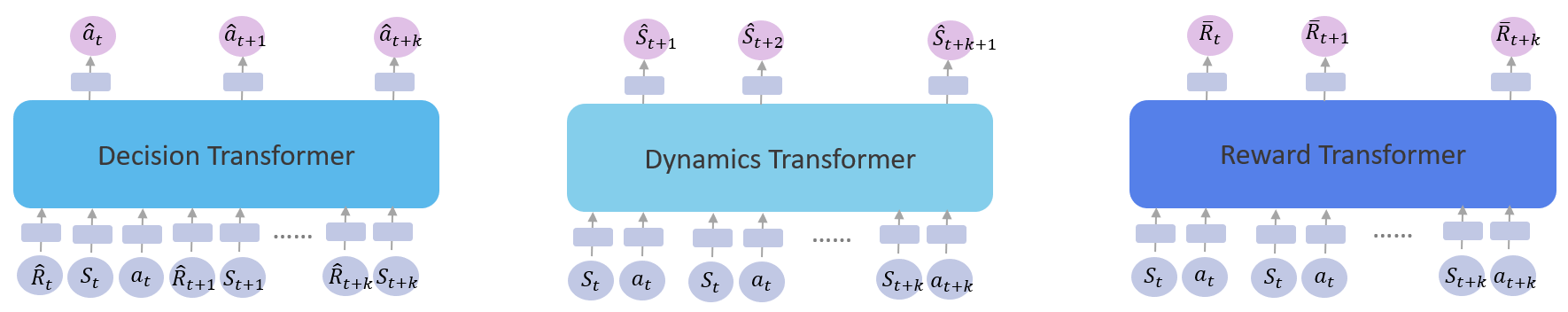}
    \vspace{-0.1in}
    \caption{Overview of the three models. We train three models in the training phase. The Decision Transformer model is trained to predict action given past trajectory. The Dynamics Transformer model is trained to predict state given past states and past actions. The Reward Transformer model is trained to predict return-to-go of the past states and actions.  }
    \vspace{-0.05in}
    \label{fig:method-threeModels}
\end{figure}

In the training phase, we train three three models over the dataset, shown in Figure \ref{fig:method-threeModels}. We feed the last K timesteps into Decision Transformer, for a total of 3K tokens. For the Dynamics Transformer and Reward Transformer, we feed last k timesteps into them but only 2k tokens, including state and actions. The output of the Dynamics Transformer is the next-timestep state and the output of reward transformer is distribution of reward-to-go value. 

\section{Experimental Results}

\subsection{Comparison on RTG models}
In this section, we investigate the performance of distributional value-estimator, comparing with the mean value-estimator. We train these distributional RTG models and mean RTG models on the same dataset of minigrid fourrooms tasks. And implement greedy planner with same dynamics transformer model to solve fourroom tasks. 

\begin{figure*}[h!]
    \centering
    \vspace{-0.2in}
    \includegraphics[width=0.9\textwidth]{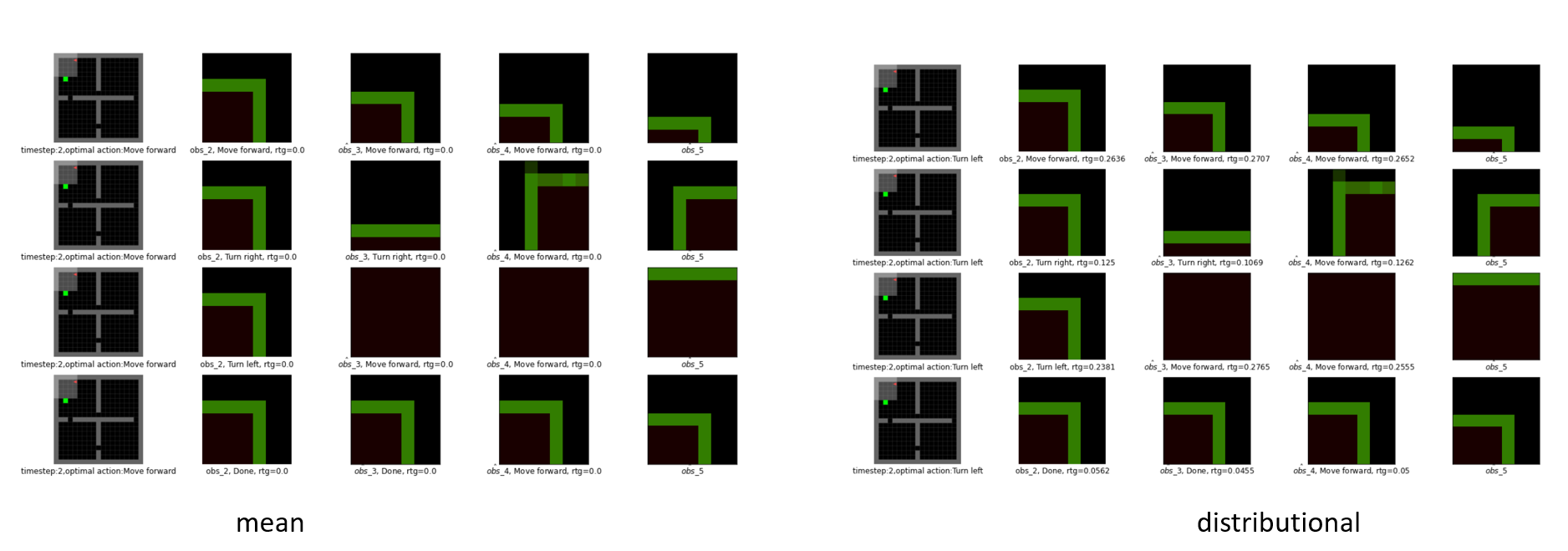}
    \vspace{-0.2in}
    \caption{Illustration of Imagined Trajectories using Tree-based Planner. This illustrates two scenarios in the planning phase. Both panels show the agent's global view and imagined trajectories based on candidate actions at one time. The text below each global-view picture shows the timestep number and optimal action based on the imagined trajectories. The text below each local-view picture shows the current trajectory's timestep number, next action, and RTG value. The left panel shows that the value estimator, which only predicts the expectation of return-to-go value. The right panel shows the distributional value estimator's evaluation. The distributional estimator performs better than  other estimators. }
    \vspace{-0.1in}
    \label{fig:rtg-vis}
\end{figure*}

From Figure \ref{fig:rtg-vis}, we can see that distributional RTG model can estimate more accurate RTG value than the mean RTG models. This is due to the fact that learning distribution of RTG value provide more accurate information about relationship between trajectory and RTG values. 

\subsection{Trajectory Imagination Using Dynamics transformer}

In this section, we investigate the performance of the Dynamics Transformer. We firstly train a dynamics model and then plan the optimal action based on imagined trajectories with optimal RTG values. In the planning phase, we imagine certain-horizon trajectories based on the candidate actions and choose the action with optimal RTG values, which is evaluated on the imaged trajectory.

\begin{figure}[h!]
    \centering
    \vspace{-0.2in}
    \includegraphics[width=0.75\textwidth]{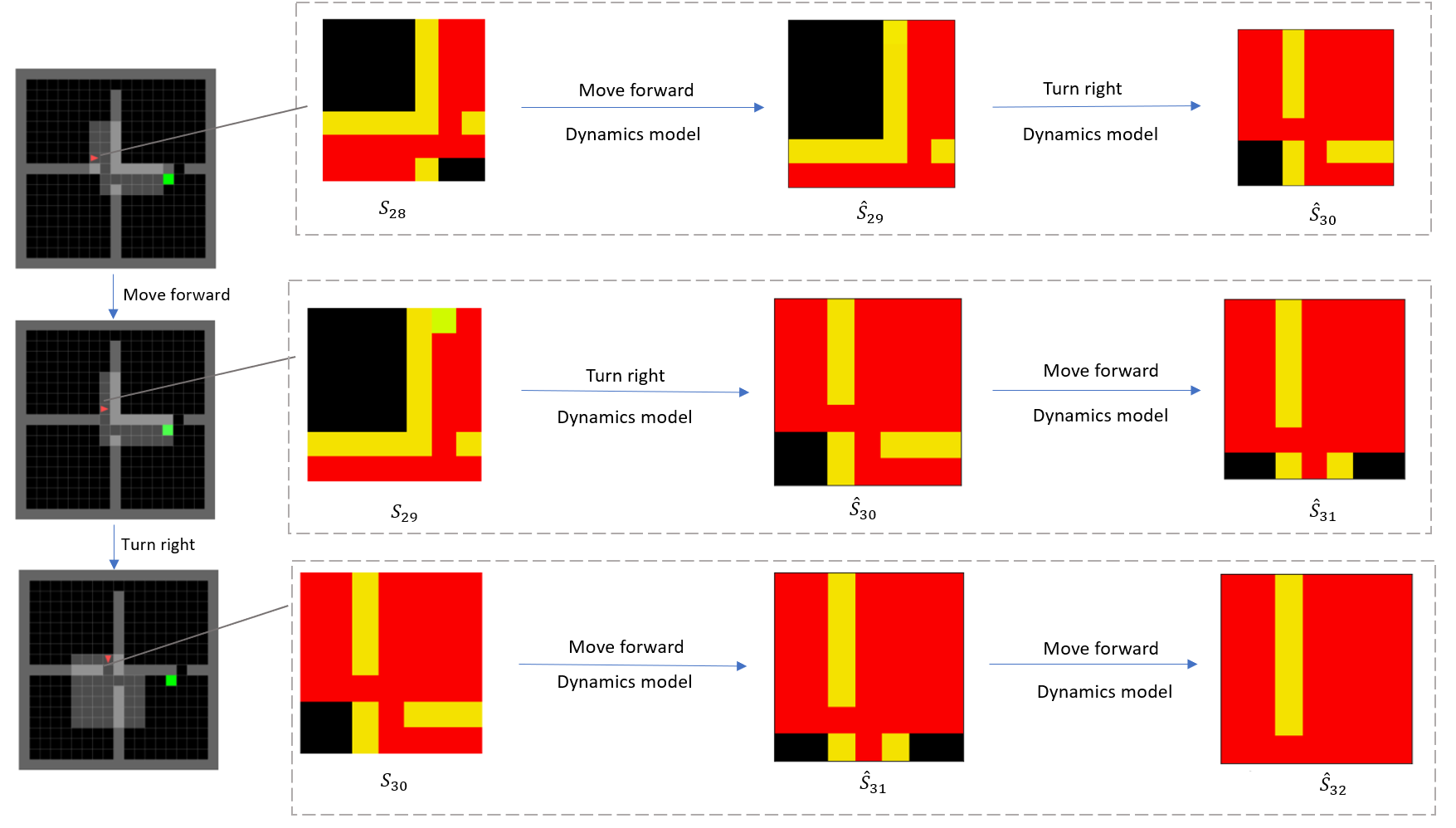}
    \vspace{-0.3in}
    \caption{Illustration of Imagined Trajectories using Dynamics Transformer. The left grid shows the agent executes action in the Fourrooms environment. The right three panels shows the imagined trajectory, starting from the true observation from the environment and predicting the next observation according to the action.  }
    \label{fig:dynamics-visualization}
\end{figure}

To demonstrate the performance of Dynamics Transformer, we visualize the imagined trajectory using the tree-based planner with the configuration, where Candidate Number=1 and Plan Horizon=2. As shown in the Figure \ref{fig:dynamics-visualization}, the Dynamics performs very well.

\subsection{Evaluation on MoEs-based models} \label{ssec:evalautionMoEs}

This section evaluates whether the MoEs layer can improve the episode reward. Here, we use the trained Decision Transformer models with MoEs layers to interact with the gym-mini-grid FourRooms environments and compared the performance. Also, we evaluate the models on the modified continuous task. In the modified setting the agent gets positive reward when moving closer to the goal instead of only getting reward when reaching the goal. This modification is a strategy of reward shaping, which is an effective technique for incorporating domain knowledge into reinforcement learning and simplifies the problem to facilitate the study of dataset collection.
In the evaluation phase, the desired target return is 1.0 in sparse-reward setting, and the desired target return is 25.0 in continuous reward setting. 

\begin{figure*}[h!]

    \centering
    \vspace{-0.1in}
    \begin{subfigure}
    [Training loss on sparse-reward setting]{    
    \centering
    \includegraphics[scale=0.5] {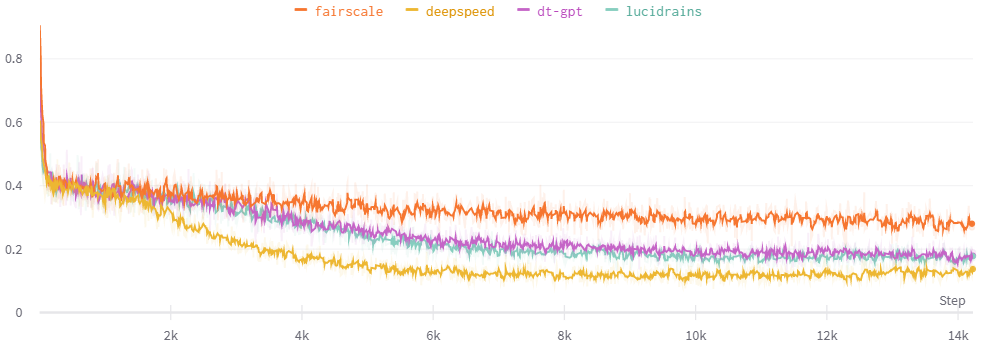}
    }
    \end{subfigure}
    \hspace{0.1in}
    \begin{subfigure}[Evaluation Reward]{
    \centering
    \includegraphics[scale=0.3] {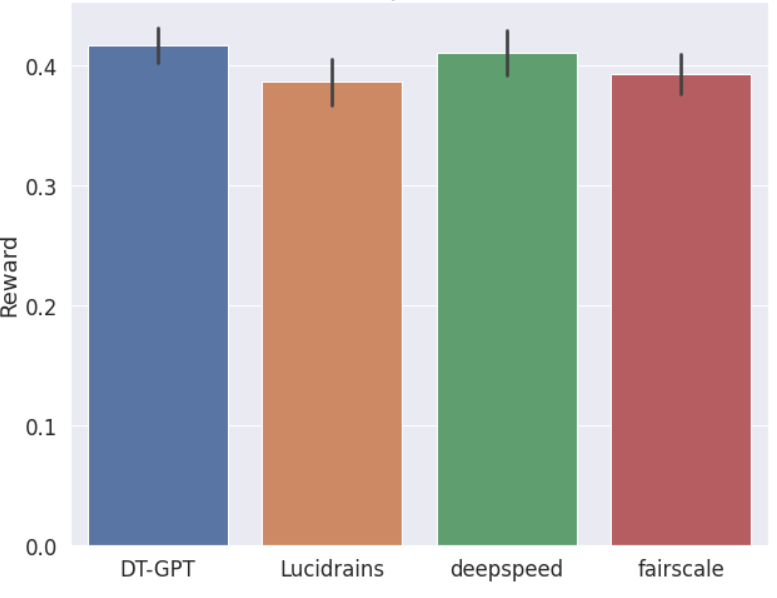}
    }
    \end{subfigure}

    \begin{subfigure}[Train Loss on continuous-reward setting]{
    \centering
    \includegraphics[scale=0.5] {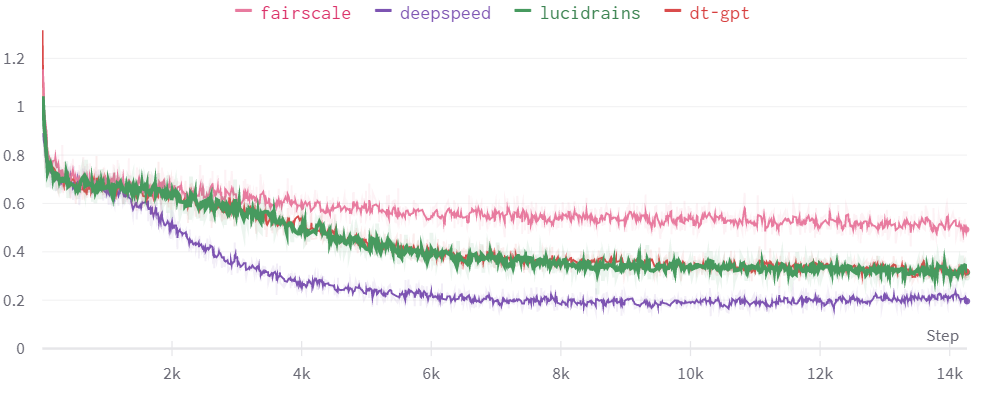}
    }
    \end{subfigure}
    \hspace{0.1in}
    \begin{subfigure}[Evaluation Reward]{
    \centering
    \includegraphics[scale=0.3] {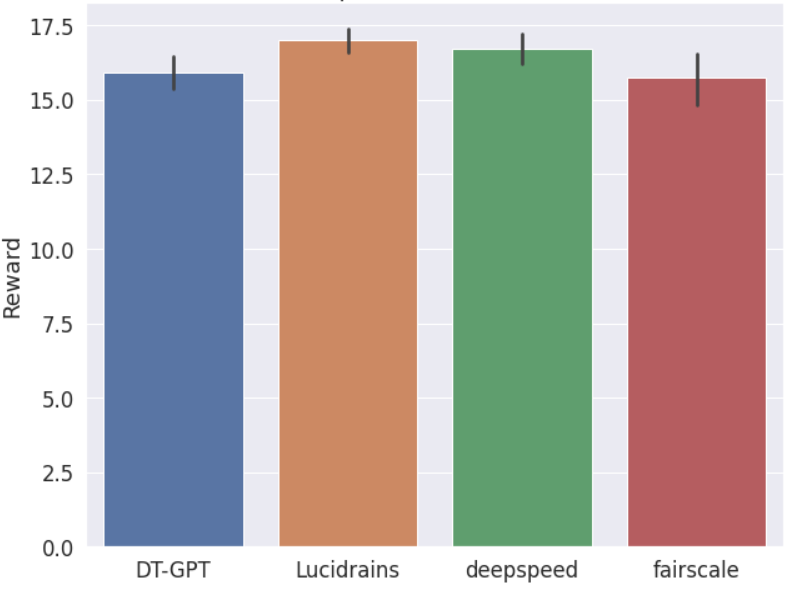}
    }
    \end{subfigure}
    
    \vspace{-0.1in}
    \caption{\small The training loss and evaluation reward in sparse-reward and continuous reward setting. The upper figures show the training loss and evaluation reward in a sparse-reward setting, while the lower figures show the continuous-reward setting. We report the training loss of each iteration in the training phase in (a) and (c). We also calculate the episode reward, averaging 1000 episode rewards over ten different seeds.}
    \vspace{-0.1in}
    \label{fig:moes-differentreward}
\end{figure*}

We plot the evaluation reward in Figure \ref{fig:moes-differentreward}. The upper figures show that the performance stay almost the same in sparse-reward setting even if the model has different training curves. This implies that better training loss on the offline dataset doesn't lead to improved evaluation reward in the sparse-reward setting. We assume this result is caused by the long-term and sparse-reward feature in FourRooms task. The lower figures show that the MoEs integrated models have marginally higher evaluation reward than the original GPT model. This implies that the MoEs with lower train loss improve the evaluation reward in the continuous reward setting.

\subsection{Overfitting on Dataset}

\begin{figure}[h!]
    \centering
    \includegraphics[width=1.0\textwidth]{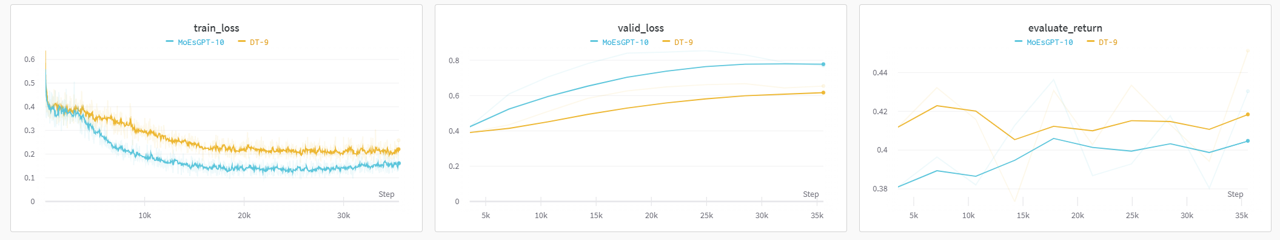}

    
    \caption{Comparison of train loss, validation loss, evaluation reward between different model architectures. The blue line represents the MoEs-based Decision Transformer, and the yellow line represents the transformer-based Decision Transformer. The training loss is collected during each iteration in the training phase of the gym-mini-grid four rooms task. The validation loss and evaluation reward are collected every epoch when training the model.
    In these figures, the MoEs-based one converges to a lower train loss than the other on the same collected dataset. But MoEs-based model shows a higher validation loss on the same validation dataset. It also has a lower evaluation reward on the four rooms task. }
    \label{fig:model-overfitting}
\end{figure}

This section investigates why the Decision Transformer (DT) model with lower train loss can not lead to higher evaluation reward. This problem is posted in the Section \ref{ssec:evalautionMoEs}. 
Since DT models the conditional distribution of actions given returns-to-go and states, lower train loss essentially is indicative that this distribution is learned well. For checking generalization during training, we use techniques similar to supervised learning. Specifically, we split the collected dataset into two parts: 80\% for the training phase and 20\% for the validation phase. Then, we keep a validation dataset of trajectories and check the validation loss is decreasing along with training loss. Meanwhile, we review the evaluation reward during the training phase. Then we plot the training loss, validation loss and evaluation reward in the Figure \ref{fig:model-overfitting}

From Figure \ref{fig:model-overfitting}, we can see that MoEs-based DT converges to lower training loss compared with transformer-based DT since the model architecture is more complex and model size is more significant. But the validation loss increases when train loss decreases. Also, MoEs-based DT goes to higher validation loss compared with transformer-based DT. This implies that the model overfits the training phase dataset and can not generalize well in the evaluation phase. The overfitting causes the model with lower train loss to perform worse in the evaluation phase. 
Also, from the subfigure (b) and subfigure (c) in the Figure \ref{fig:model-overfitting}, we can conclude that the model with lower validation can generalize better in the evaluation phase. To get higher validation loss and better performance in learning the task, we decrease the training epochs and reduced the model size in our experiment. 






\end{document}